\documentclass[11pt]{article}
\usepackage{coling2020}
\usepackage{times}
\usepackage{url}
\usepackage{latexsym}

\newcommand{\ACRO}[1]{\textsc{#1}}
\newcommand{\ARRAU}{\ACRO{arrau}}
\newcommand{\BERT}{\ACRO{bert}}
\newcommand{\BRIDGING}{\ACRO{element-of}}
\newcommand{\COREF}{\ACRO{single-coref}}

\newcommand{\CONCAT}{\ACRO{concat}}
\newcommand{\LINGEX}[1]{\textit{#1}}

\newcommand{\ONTONOTES}{\ACRO{ontonotes}}
\newcommand{\PD}{\ACRO{pd}}
\newcommand{\PDGAME}{\textit{Phrase Detectives}}
\newcommand{\PDSILVER}{\ACRO{pd-silver}}
\newcommand{\PDCROWD}{\ACRO{pd-crowd}}
\newcommand{\PRED}[1]{\textbf{#1}}
\newcommand{\PRETRAIN}{\ACRO{pre-train}}
\newcommand{\TEACHER}{\ACRO{annealing}}

\newenvironment{EXAMPLE}{\begin{list}{}
    {\topsep      5pt
     \itemsep     .0ex
     \labelwidth  25pt
     \leftmargin  30pt
     
     }}{\end{list}}
\newcommand{\ENEW}[1]{\refstepcounter{equation}
\label{#1}\item[(\theequation)\hspace{10pt}]}
\newcommand{\EITEM}{\stepcounter{EEXAMPLE}\item[\theEEXAMPLE.]}
\newcounter{EEXAMPLE}
\renewcommand{\theEEXAMPLE}{\alph{EEXAMPLE}}

\newenvironment{AEXAMPLE}{\begin{list}{}
    {\topsep      0pt
     \partopsep   0pt
     \itemsep     .0ex
     \labelwidth  2pt
     \leftmargin  0pt
     
     \usecounter{EEXAMPLE}
     }\vskip-\lastskip}{\end{list}}
\newcommand{\ENUMA}[1]{\refstepcounter{equation}
\label{#1}\item[(\theequation)] \begin{AEXAMPLE}}
\newcommand{\ENDENUMA}{\end{AEXAMPLE}}
\newcommand{\SREF}[1]{(\ref{#1})}
\newcommand{\SREFA}[2]{(\ref{#1}#2)}

\usepackage{amsmath}
\usepackage{booktabs}
\usepackage{tabularx}
\usepackage{footnote}

\usepackage{multirow}
\usepackage{graphicx}
\usepackage{subcaption}
\usepackage[dvipsnames]{xcolor}
\definecolor{cMISSING}{gray}{0.7}
\colorlet{cWRONG}{Red!60!}
\colorlet{cNESTWRONG}{Red!100!}
\colorlet{cCORRECT}{Green!40!}

\newcommand{\MISSING}[1]{\colorbox{cMISSING}{\textbf{#1}}}
\newcommand{\WRONG}[1]{\colorbox{cWRONG}{\textbf{#1}}}
\newcommand{\NESTWRONG}[1]{\colorbox{cNESTWRONG}{\textbf{#1}}}
\newcommand{\CORRECT}[1]{\colorbox{cCORRECT}{\textbf{#1}}}
\newcommand{\ANAPHOR}[1]{\underline{\textbf{#1}}}

\colingfinalcopy % Uncomment this line for the final submission

\title{Free the Plural: Unrestricted Split-Antecedent Anaphora Resolution}

\author{Juntao Yu$^1$, Nafise Sadat Moosavi$^2$, Silviu Paun$^1$, Massimo Poesio$^1$ \\
  $^1$Queen Mary University of London \\
  $^2$UKP Lab, Technische Universität Darmstadt\\
  $^1${\tt \{juntao.yu, s.paun, m.poesio\}@qmul.ac.uk} \\
  $^2${\tt moosavi@ukp.informatik.tu-darmstadt.de}}

\date{}

\begin{document}
\maketitle
\begin{abstract}
Now that the performance of coreference resolvers on  the simpler forms of anaphoric reference has greatly improved,
more attention is  devoted to more complex aspects of anaphora.
% in recent years, but the scope of most coreference resolvers is still pretty limited.
% The performance of coreference resolution systems has greatly improved in recent years, but the scope of most coreference resolvers is still pretty limited.
One limitation of virtually all coreference resolution models
is %that 
%has been largely improved during the recent years. System developed for coreference resolution 
%coreference resolution models only 
the focus on  single-antecedent anaphors. 
Plural anaphors with multiple antecedents--so-called split-antecedent anaphors (as in \textit{John met Mary. \underline{They} went to the movies})--have not been widely studied, %in {\NLP},
because they are not annotated in {\ONTONOTES} and are relatively infrequent in other corpora.
In this paper, we introduce the first model for
unrestricted resolution %that is able to resolve unrestricted 
of split-antecedent anaphors. 
We start with a strong baseline enhanced by {\BERT} embeddings, 
and show that 
we can substantially improve its performance
%its performance can be substantially improved 
by addressing the sparsity issue. 
%largely 
To do this, we experiment with  auxiliary corpora 
%created from the crowd annotated 
where split-antecedent anaphors were annotated by the crowd, 
and with transfer learning models using 
 \PRED{element-of} bridging references 
and  single-antecedent coreference as auxiliary tasks. 
%using three different training strategies. 
Evaluation on the gold annotated {\ARRAU} corpus shows that the out best model uses %best results are obtained using 
%auxiliary corpus created from 
%single-antecedent coreference as an auxiliary task;
a combination of three auxiliary corpora
achieved F1 scores of 70\% and 43.6\% when evaluated in a lenient and strict setting, respectively, i.e., 11  and 21 percentage points gain when compared with our %strong 
baseline.\footnote{The code is available at %\url{http://AnonymousLink}
\url{https://github.com/juntaoy/dali-plural}
}
% scores of 
% %69.6\% and 42.7\% 
% 70\% and 43.6\% 
% were obtained when evaluated in a lenient and strict setting respectively. 
% This is an 11\% and 21\% gain when compared with our strong baseline.
%in a lenient and strict evaluation respectively.
\end{abstract}

\section{Introduction}
\textbf{(Identity) anaphora resolution} (\textbf{coreference})
is the linguistic task of linking nominal expressions (mentions) to entities in the discourse, so that  mentions representing the same entity are grouped together in a `coreference chain' \cite{poesio-stuckardt-versley:book}. 
% It is important to downstream applications that require deep understanding on the context, such as summarization \cite{steinberger-et-al:IPM07,steinberger-et-al:ana_book_summarization}. 
As the performance of  coreference 
models %system 
has 
substantially %largely 
improved \cite{clark2015entity,clark2016improving,lee2017end,lee2018higher,kantor-globerson-2019-coreference} in recent years,
more attention is being devoted to more complex aspects of anaphoric reference--
from the %cases of 
pronouns that require commonsense knowledge for their resolution studied in the Winograd Schema Challenge \cite{rahman&ng:EMNLP2012,peng-et-al:NAACL2015}
to pronouns that cannot be resolved purely on the basis of gender \cite{webster-et-al:TACL2018}.
Another limitation of state of the art systems is the assumption that 
%coreferent 
anaphors can only have one antecedent. 
%The more complex cases in which 
Plural anaphors with multiple antecedents (\textbf{split antecedent anaphors}) are not widely studied-- 
in fact, 
such %the split antecedent 
anaphors are not annotated in the most widely used coreference corpus, {\ONTONOTES}  \cite{pradhan2012conllst}.
In {\ONTONOTES} we find annotated cases of plural reference to plural antecedents, as in \SREF{ex:simple}, 
or cases in which singular antecedents are conjoined
so that a mention can be introduced for the conjunction,
as in \SREF{ex:conjunction}. 
% Some of the simpler plural cases that mentions are linked by conjunctions were annotated in the OntoNotes corpus by introducing a new mention, e.g. in
% \SREF{ex:simple}. 
However, it is also possible to refer plurally to antecedents introduced by separate noun phrases, as 
in %the harder cases such as in 
\SREF{ex:hard} or \SREF{ex:common} \cite{eschenbach1989remarks,kamp&reyle:93}; such cases are not annotated in  {\ONTONOTES}. 

\begin{EXAMPLE}
\ENEW{ex:simple} 
\textit{The Joneses}$_i$ went to the park. \underline{They}$_i$ had a good time.
\ENEW{ex:conjunction} 
\textit{John and Mary}$_i$  went to the park.  \underline{They}$_i$  had a good time.
\ENEW{ex:hard}
\textit{John}$_i$ met \textit{Mary}$_j$ in the park. \underline{They}$_{i,j}$ had a good chat .
\ENEW{ex:common}
John likes \textit{green}$_i$, Mary likes \textit{blue}$_j$, but Tom likes \underline{both colours}$_{i,j}$. 
\end{EXAMPLE}

Early research on split antecedent anaphors 
mostly focused on the constraints on 
%are mainly focused on the linguistic discussions about 
the construction of complex entities from singular entities
%how complex entities are constructed from singular entities \cite{eschenbach1989remarks,kamp&reyle:93}.
There are two recent studies of split antecedent anaphora,
both involving the creation of a new dataset, and focusing on a subset of the problem. 
\newcite{vala-etal-2016-antecedents} proposed a model focused on the resolution of the 
%is the only recent system that tried to resolve split-antecedent anaphors. Their system is designed 
%to solve two 
pronominal plural mentions \textit{they} and \textit{them} %``they" and ``them''
in a newly created corpus of fiction 
%fiction domain corpus 
where most of the antecedents are from a fixed list of 
%countable 
characters in the novel. 
\newcite{zhou&choi:COLING2018} proposed a model that addresses a wider range of anaphoric references to split antecedents, but limited only to references to main characters (mainly pronominal mentions) in a corpus of transcripts of the sitcom \textit{Friends}.

In this paper, we introduce the first model targeting the whole range of split-antecedent anaphora.
We evaluate our system on the 
hand-annotated %gold annotated 
{\ARRAU} corpus \cite{poesio_anaphoric_2008,uryupina-et-al:NLEJ},
which covers a range of anaphoric relations going from 
identity relations (including split antecedent anaphora), 
bridging reference, and discourse deixis,
%The {\ARRAU} corpus has a rich list of anaphoric relations annotated such as coreference relations (including split-antecedent anaphors), bridging references, discourse deixis which well suits our task.
as well as different genres going from news to task-oriented dialogue.
Since the task is complex, 
we focus on establishing links between (gold) plural anaphors and their split-antecedents,
leaving the detection of  split-antecedent anaphors  for future work. 
We follow \newcite{vala-etal-2016-antecedents} and evaluate on the setting that assumes the gold split-antecedent anaphors and the gold mentions are provided.

Our baseline system is a simplified version of the state of the art coreference system \cite{lee2018higher,kantor-globerson-2019-coreference} enhanced by {\BERT} embeddings \cite{devlin2019bert}.
The key issue we tackled is that compared to single-antecedent anaphors, the number of split-antecedent anaphors is rather small: in total,  only 697 split-antecedent anaphors were annotated in the {\ARRAU} corpus, about 2\% of all anaphoric references. 
To tackle this challenge of 
limited training data, %lacking the training data, 
we experimented with different ways  of 
\textbf{using auxiliary corpora} to improve  performance. 
We evaluated four different augmentation settings.
Two of these involve using additional examples of split antecedent anaphora recoverable 
%for the first two settings, we create different
%creating auxiliary corpora 
from the crowdsourced {\PDGAME}  corpus ({\PD}) \cite{poesio-etal-2019-crowdsourced},
%using different aggregating methods. 
a corpus %The {\PD} corpus consists 
of  anaphoric annotations, including split-antecedent anaphors, collected using the {\PDGAME}  game.\footnote{\url{https://www.phrasedetectives.org/}}
The corpus includes %comes with 
both raw annotations and 
%as well as 
silver labels aggregated using the Mention Pair Annotations model \cite{paun-etal-2018-probabilistic}. 
For our first setting ({\PDSILVER}), we used the silver labels to identify split-antecedent anaphors in the corpus, and use them as an auxiliary corpus. 
For the second setting ({\PDCROWD}), we use the same data as {\PDSILVER}, but instead of using silver labels we collect all  `raw' split-antecedent annotations, and use majority voting to choose the labels when there are different annotations for the same anaphor. 
The third and fourth 
approaches to augmentation %settings 
involve %using 
a form of transfer learning, using annotations of different but related phenomena to help learning split antecedent resolution.
In our third setting ({\BRIDGING}), we used as auxiliary training data examples of a type of  bridging reference, \PRED{element-of},  that is related  to our task and is annotated in the {\ARRAU} corpus.\footnote{The relation between the split-antecedent anaphors and the individual antecedents is the inverse of \PRED{element-of}.}
Finally, in our last setting ({\COREF}), we used as auxiliary training data the examples of
%a relatively large auxiliary corpus using 
 single-antecedent coreference  annotated in the {\ARRAU} corpus.

We also evaluated three different training strategies to 
leverage %integrate 
the auxiliary data.
%the auxiliary corpora most effectively. 
The first strategy ({\CONCAT}) involves  randomly selecting a training document from the main corpus and the auxiliary corpus in turns, with a fixed probability of 0.5 to train with the main corpus. 
For the second strategy ({\PRETRAIN}), we first pre-train our model on the auxiliary corpus, and fine-tune it on the main corpus. 
Our last strategy ({\TEACHER}) is inspired by the teacher annealing method used in \newcite{clark-etal-2019-bam}.
We train our models with both the main corpus and the auxiliary corpus as in the {\CONCAT} strategy, but linearly increase the usage of documents from the main corpus. In this way, the training progressively switches from the auxiliary corpus  to the main corpus. 

The evaluation on the {\ARRAU} corpus shows that all four auxiliary datasets combined with all three training strategy substantially improve the performance of the baseline model. The evaluation on mixed use of our auxiliary corpora results in further improvements on performance.
Our best model, trained with three auxiliary corpora ({\PDCROWD}, {\BRIDGING}, {\COREF}), outperforms our strong baseline by 11.4  and 20.9 percentage points when evaluated in a lenient and strict setting, respectively. 
The final model has an F1 score of 70\% when partial credit is granted in the evaluation, and correctly resolves all antecedents for 43.6\% of the split-antecedent anaphors (the strict evaluation). 
To the best of our knowledge, this is the first reported result on unrestricted split-antecedent anaphora resolution.

\section{Background}
\subsection{Single-Antecedent Coreference Resolution}

Single-antecedent 
coreference resolution
has been extensively studied. 
In the pre-neural period, both rule-based \cite{lee-et-al:CL13} and statistical %feature-based 
models \cite{soon-et-al:CL01,bjorkelund2014learning,clark2015entity} 
were developed %have been developed 
to resolve %to resolving 
single-antecedent coreference resolution. 
Recently, \newcite{wiseman2015learning,wiseman2016learning} first introduced a neural network-based approach to solving coreference in a non-linear way. 
\newcite{clark2016improving} integrated reinforcement learning to let the model, optimising directly on the B$^3$ scores. 
\newcite{lee2017end}  proposed a neural joint approach for mention detection and coreference resolution. Their model does not rely on parse trees; instead, the system learns to detect mentions by exploring the outputs of a \ACRO{bilstm}. 
After the introduction of  contextual word embeddings such as \ACRO{elmo} \cite{peters2018elmo} and {\BERT} \cite{devlin2019bert}, the \newcite{lee2017end} system has been greatly improved by those embeddings \cite{lee2018higher,kantor-globerson-2019-coreference,joshi2019bert,joshi2019spanbert} to achieve SoTA results. But none of those SoTA systems can resolve split-antecedent anaphors. 

\subsection{Split-Antecedent Anaphora}

% I don't think it's necessary
%\textbf{I am going to add here a discussion of the phenomenon with examples from Eschenbach et al }
There is  substantial research on resolving split-antecedent anaphors both in linguistics \cite{kamp&reyle:93} and  psychology \cite{murphy:84,sanford&lockhart:JOS1990,kaup-et-al:LCP2002,patson:compass2014},
but only a few early computational models
%are mainly focused on linguistic level
\cite{eschenbach1989remarks}.
This work was primarily concerned with explaining preferences and restrictions on split antecedent anaphora: for example, in \SREFA{ex:split-constraints}{a} \LINGEX{they} can refer to Michael, Peter and Maria, or to Peter and Maria, but not to Michael and Maria;
or in \SREFA{ex:split-constraints}{b} there seems to be a preference  for \LINGEX{they} to refer to Peter and Maria \cite{kaup-et-al:LCP2002}.
Proposals differed, e.g., on whether complex reference objects are created immediately or after encountering the anaphor.
%, and the difference between   \LINGEX{they} and \LINGEX{both}.

\begin{EXAMPLE}
\ENUMA{ex:split-constraints}
\EITEM Michael met Peter and Maria in the pub. They had a great time.
\EITEM Michael watched Peter and Maria in the pub. They had a great time.
\ENDENUMA
\end{EXAMPLE}
% These proposals  focused on issues such as whether  complex reference objects are created immediately or on demand, and on the differences between the use of  \LINGEX{they} and \LINGEX{both} \cite{murphy:84,sanford&lockhart:JOS1990,kaup-et-al:LCP2002,patson:compass2014}.
Only two recent computational treatments of this type of anaphor exist \cite{vala-etal-2016-antecedents,zhou&choi:COLING2018};
%in this area exist, 
they are discussed in Section \ref{section:relatedwork}. %Related Work section. %the section is not called related work

\subsection{Approaches for Under-Resourced Tasks}
Much research has focused on resolving under-resourced tasks via semi-supervised learning \cite{pekar-etal-2014-exploring,yu-bohnet-2015-exploring,kocijan-etal-2019-wikicrem,hou-2020-acl} and shared representation based transfer learning \cite{yang2017transfer,cotterell2017low,zhou2019dual}.

\paragraph{Semi-supervised learning}
Semi-supervised methods use large unlabelled/automatically labelled data to enhance %the system 
performance for under-resourced domains/languages. Early research %was mainly 
focused on generating additional training data from  automatically annotated text. \newcite{pekar-etal-2014-exploring}, \newcite{yu-etal-2015-domain} used co-training/self-training for dependency parsing to leverage models trained on  rich-resourced domain for under-resourced domains.  \newcite{yu-bohnet-2015-exploring} applied a confidence-based self-training approach to enhance parsing performance for 9 languages with parse trees automatically annotated by models trained on small initial training data.  

Recently, another line of research
focused on creating %starts to create 
synthetic training data from unlabelled/automatically labelled data using some heuristic patterns. 
\newcite{kocijan-etal-2019-wikicrem} used  Wikipedia to create WikiCREM, a large  pronoun resolution dataset,  using 
%a small number of 
heuristic rules based on the occurrence of personal names in  sentences. The evaluation of their system on  Winograd Schema corpora shows that  models pre-trained on the WikiCREM consistently outperform  models that do not  use  it.
\newcite{hou-2020-acl} applied a similar approach to the  antecedent selection task for bridging references. 
Hou created an artificial  bridging corpus using the prepositional and 
possessive structures in the automatically parsed  Gigaword corpus. 
Models pre-trained with the artificial corpus achieved substantial gains 
over %when compared with the 
baselines. 
Our {\PDSILVER} and {\PDCROWD} settings are close to this approach, but instead of using automatically annotated data, we use crowdsourced annotations from the {\PD} corpus. 
Both our corpora and the synthetic corpora created by \newcite{hou-2020-acl} contain 
a degree % certain degrees 
of noise  compared with gold annotated corpora.

\paragraph{Shared Representation-Based Transfer Learning}
Shared representation-based transfer learning focuses on exploiting auxiliary tasks 
for which large annotated data exist 
to help in
%(usually rich in resources) to aid the
under-resourced tasks/domains/languages. It is similar to multi-task learning, but only focuses on enhancing the performance of the under-resourced task.
\newcite{yang2017transfer} applied transfer learning to sequence labelling tasks; the deep hierarchical
recurrent neural network used in their work is fully/partially shared between the source and the target tasks. 
They demonstrated that SoTA performance can be achieved by using models trained on multi-tasks. 
\newcite{cotterell2017low} trained a neural NER system on a combination of high-/low-resource languages to improve  NER for the low-resource languages. In their work,  character-based embeddings are shared across the languages. Recently, \newcite{zhou2019dual} introduced a multi-task network together with adversarial learning for under-resourced NER. The evaluation on both cross-language and cross-domain settings shows that partially sharing the \ACRO{bilstm} works better for cross-language transfer, while for cross-domain setting, the system performs better when the \ACRO{lstm} layers are fully shared. Our third and fourth settings ({\BRIDGING} and {\COREF}) can be viewed as shared representation-based transfer learning where we use bridging resolution and  single-antecedent coreference resolution as our auxiliary tasks to aid our split-antecedent anaphora resolution.

\section{Methods}
\subsection{The Baseline System}

Our baseline is %For our baseline model, we use 
a simplified version of the SoTA coreference architecture by \newcite{lee2018higher}, 
further developed by \newcite{kantor-globerson-2019-coreference}.
%The \newcite{kantor-globerson-2019-coreference} model is an extended version of  \newcite{lee2018higher}; the main difference is that 
%Kantor et al use
%{\BERT} embeddings \cite{devlin2019bert} instead of the ElMo embeddings \cite{peters2018elmo} used by
%Lee et al. 
%These systems have similar architecture and both
In this model, %The system do 
mention detection and coreference are carried out jointly,
but here %in this paper 
we only use the coreference part,
%of the system, 
since %for our task 
we evaluate our model with gold mentions. 

Our baseline system first creates representations for mentions using the output of a \ACRO{bilstm}. 
%The sentences of a document are encoded from both directions to obtain a representation for each token in the sentence. 
The \ACRO{bilstm} takes as input the concatenated embeddings 
%($(emb_t)_{t=1}^{T}$) 
of both word and character levels.
For word embeddings,  GloVe \cite{pennington2014glove} and {\BERT} \cite{devlin2019bert} embeddings are used. 
Character embeddings are learned from a convolution neural network (CNN) during training. 
The tokens are represented by concatenating outputs from the forward and the backward \ACRO{lstm}s. 
The token representations $(x_t)_{t=1}^{T}$  are used together with head representations ($h_i$) to represent mentions ($M_i$). 
The $h_i$ of a mention is obtained by applying 
attention over its token representations ($\{x_{b_i}, ..., x_{e_i}\}$), where $b_i$ and $e_i$ are the indices of the start and the end of the mention, respectively. 
Formally, we compute $h_i$, $M_i$ as follows:

\vspace{-10pt}
% \small
$$\alpha_t = \textsc{ffnn}_{\alpha}(x_{t})$$
$$a_{i,t} = \frac{exp(\alpha_t)}{\sum^{e_i}_{k=b_i} exp(\alpha_k)}$$
$$h_i = \sum^{e_i}_{t=b_i} a_{i,t} \cdot x_t $$
$$M_i = [x_{b_i}, x_{e_i},h_i,\phi(i)]$$
% \normalsize
where $\phi(i)$ is the mention width feature embeddings. 
Next, we pair the mentions with candidate antecedents to create a pair representation 
($P_{(i,j)}$):

\vspace{-8pt}
% \small
$$P_{(i,j)} = [M_i, M_j, M_i \circ M_j,\phi(i,j)]$$
% \normalsize
where $M_i$, $M_j$ is the representation of the antecedent and anaphor, respectively, 
$\circ$ denotes  element-wise product, and
$\phi(i,j)$ is the distance feature between a mention pair. 
To make the model computationally tractable, we consider for each mention a maximum 250 candidate antecedents as we observed from the {\ARRAU} corpus, most of the antecedents can be retrieved within the 250 candidates window size.

The next step is to compute the pairwise score ($s(i,j)$). Following \newcite{lee2018higher}, we add an artificial antecedent $\epsilon$ to deal with cases of non-split-antecedent anaphor mentions or cases when the antecedent does not appear in the candidate list during the training. We do not use $\epsilon$ during test time as we use gold split-antecedent anaphors. We compute $s(i,j)$ as follows:

\vspace{-8pt}
%\small
$$r(i,j) = \textsc{ffnn}(P_{(i,j)})$$
$$s(i,j) = \frac{1}{1+e^{-r(i,j)}} $$
% \normalsize
%
At %During the 
test time, the system will generate two to five antecedents according to their $s(i,j)$ scores. The upper threshold is based on the observation that the vast majority of  split-antecedent anaphors in  {\ARRAU}  have no more than 5 antecedents. 
To generate the antecedents, we first rank the candidates by their $s(i,j)$ scores in descending order. We add up to 5 top candidates that have a $s(i,j)$ score above 0.5.\footnote{We use the gold single-antecedent clusters to ensure the selected antecedents belong to distinct gold cluster.} 
If %In the cases, 
less than two candidates were selected, we add top two candidates to the predictions regardless of their scores. %Figure \ref{fig:nngraph} shows the proposed architecture of our system.

% \begin{figure}[t]
%     \centering
%     \includegraphics[width=.7\columnwidth]{nn-graph.pdf}
%     \caption{The proposed architecture of our system.
%     }
%     \label{fig:nngraph}
% \end{figure}

\subsection{Auxiliary Corpora}
\label{sec:auxiliary}

Since the number of examples of split-antecedent anaphora in {\ARRAU} is %rather 
small, we deployed four auxiliary corpora created from either the crowd annotated {\PDGAME} ({\PD}) corpus or the gold annotated {\ARRAU} corpus to improve the performance of the system. 

The {\PD} corpus was created using the {\PDGAME} game, 
whose %by asking 
players are asked to find the antecedent/split-antecedents %that are 
closest to the mention in question \cite{poesio-etal-2019-crowdsourced}.
The corpus comes with all raw annotations and silver labels aggregated using the Mention-Pair Annotation model \cite{paun-etal-2018-probabilistic}. We created our first two auxiliary corpora from the latest version of the {\PD} corpus\footnote{\url{https://github.com/dali-ambiguity/Phrase-Detectives-Corpus-2.1.4}} by using different aggregation methods. The {\ARRAU} corpus consists of texts from four very distinct domains: news (the \ACRO{rst} subcorpus), dialogue (the \ACRO{trains} subcorpus),  fiction (the \ACRO{pear} stories), and medical / art history (the \ACRO{gnome} subcorpus). 
Its annotation scheme covers the annotation of referring (including singletons) and non-referring expressions; 
coreference relations including split antecedent plurals and generic references; 
and non-coreferential anaphoric relations including discourse deixis and bridging references. 
We create the other two auxiliary corpora from the {\ARRAU} corpus. The rest of the subsection describes our auxiliary corpora in detail.\footnote{The ParCorFull corpus \cite{Lapshinova-Koltunski:2018} also includes split-antecedent annotations, but the number of split-antecedent examples are too small to be used in our experiments.}

\textbf{Silver Labels ({\PDSILVER})} 
For our first auxiliary corpus, we 
simply added to our training data
%use a straightforward way to enhance our training data by adding 
the split-antecedent anaphora examples from the {\PD} corpus. 
We used the silver labels that come with the corpus and extracted 507 split-antecedent anaphors (see Table \ref{table:corpus_stat}). 
This nearly doubled the size of our training data. We assessed the quality of the silver labels by comparing it against the gold annotated subset of the {\PD} corpus;\footnote{A subset of the {\PD} corpus comes with additional gold labels annotated by experts.} the silver labels have a relatively good quality (62.9\% F1), recalling 68.8\% of the gold split-antecedent anaphoric links and with a precision of 57.9\%.

\textbf{Raw Crowd Annotations ({\PDCROWD})} 
The second auxiliary corpus was created by extracting all split-antecedent examples from the raw annotations in {\PD} to maximise  recall. 
After  extracting all  split-antecedent annotations,
%from the corpus and use 
we used majority voting as our aggregation method when players did not agree on split-antecedent annotations. 
In this way, we extracted 47.7k split-antecedent annotations associated with 6.2k mentions (Table \ref{table:corpus_stat}). 
The quality 
of this extraction method %our corpus 
was evaluated on the gold portion of the {\PD} corpus as well; the resulting dataset has a recall of 91.7\%, which fulfils the goal of this setting. 
As expected, the corpus is noisy, with a precision  of 11.1\% and an F1 of 19.7\%. 
We manually checked the false-positive examples, 
finding they are mainly due to 
three types of mistakes: 
single-antecedent coreference (the coreference chains were annotated as the split-antecedent), 
bridging reference (not required to be annotated),
and other annotation mistakes. The first two types of mistakes are not harmful to our task as our third and fourth auxiliary corpora are created using those types of relations.

\begin{table}[t]
\centering
%\small
\resizebox{\textwidth}!{
\begin{tabular}{lllcc}
\toprule
\bf Corpus&\bf Anaphors Type&\bf Data Quality&\bf Num of docs&\bf Num of Anaphors\\\midrule
{\ARRAU} {\ACRO{train}}/{\ACRO{dev}}/{\ACRO{test}}&Split-antecedent anaphors&Gold&211 / 30 / 60&507 %(908)$^*$ 
/ 80 / 110\\ 
%{\ARRAU} {\ACRO{dev}}&Split-antecedent anaphors&Gold&30&80\\ 
%{\ARRAU} {\ACRO{test}}&Split-antecedent anaphors&Gold&60&110\\
{\PDSILVER}&Split-antecedent anaphors&Silver&165&507\\
{\PDCROWD}&Split-antecedent anaphors&Noisy&467&6262\\
{\BRIDGING}&Bridging anaphors&Gold&213&1059\\
{\COREF}&Single-antecedent anaphors&Gold&462&30372\\
\bottomrule
\end{tabular}}
\caption{\label{table:corpus_stat} Statistics about the corpora used in our experiments.} %* indicate the number after extend the split-antecedent anaphors by their siblings in the same single-antecedent cluster.}
\end{table}

\textbf{Element-of Bridging References ({\BRIDGING})} 
{\ARRAU} is also annotated with bridging references, and one of the bridging relations covered by the annotation, \PRED{element-of} (and its inverse)
are %is one of the bridging relations 
very closely related to the task of resolving split-antecedent plurals.
\PRED{Element-of}  is the relation between a new singular entity and a plural entity introduced in the discourse, as in \SREFA{ex:element-of}{a},
or between a previously introduced singular entity and a new plural entity, 
as in \SREFA{ex:element-of}{b}.

\begin{EXAMPLE}
\ENUMA{ex:element-of}
\EITEM There are \textit{two supermarkets} in our village, but \underline{one} is very small. (\textbf{element-of})
\EITEM Yet \textit{another small bookshop} just opened in our village.
       \underline{Our independent  bookshops} are our main attraction.
       (\textbf{element-of-inverse})
\ENDENUMA
\end{EXAMPLE}
Since the proposed system uses a pairwise approach, the relations between  split-antecedent anaphors and their antecedents are established by multi-links between anaphors and individual antecedents. 
These are \PRED{element-of} relations, but 
differ from the bridging case in two respects.
%with two major differences: 
First, the plural relations are an inverse version of the element-of relations where the antecedent is an element of the anaphor.
Second,  split-antecedent coreference \emph{is} coreference:
the union of all antecedents has the same denotation as the anaphor, 
unlike in bridging.
%but for bridging there is no such constrain. 
Nevertheless, %we hypothesise that 
the \PRED{element-of} bridging relation is close enough to be potentially useful for our task. 
We therefore created 
a %Hence we create our 
third auxiliary corpus by extracting \PRED{element-of} bridging relations from  {\ARRAU}. 
In total we extracted 1059 training examples (see Table \ref{table:corpus_stat}).

\textbf{Single-antecedent anaphors ({\COREF})} 
%Finally we create 
Our last auxiliary corpus using single-antecedent anaphors. 
The main reason 
for using %that we choose to use 
single-antecedent anaphors 
as supporting dataset %to form our auxiliary corpus 
is that  single-antecedent anaphors are very common:
e.g., in {\ARRAU} {\ACRO{train}} we only have 500 split-antecedent anaphors, but 30k single-antecedent anaphors (see Table \ref{table:corpus_stat}). 
This gives us a much larger corpus than all other auxiliary corpora proposed earlier. 
Using a large auxiliary corpus allows our system to learn a better mention and pairwise representations that might be beneficial for our under-resourced task.

\subsection{Training Strategies}
Training with multiple corpora is challenging, especially when the auxiliary corpus is noisy. In this paper, we evaluate our system with three different training strategies to maximise the performance on split-antecedent anaphora resolution.

\textbf{Concatenation ({\CONCAT})} 
The first and simplest strategy is to %For the first strategy, we 
use the auxiliary corpus as additional training data by concatenating it with the main corpus. We configured  training to train on documents from the main and auxiliary corpus in turn,  with 50\% of the time on the main corpus. By doing so, we make sure the system will not overfit the auxiliary corpus.

\textbf{Pre-training ({\PRETRAIN})} Our second strategy was to first pre-train the system on the auxiliary corpus,
%to provide a good base for the main task,
and then 
fine-tuning the model %We then fine-tuning the system 
on the main corpus to fit our task. 
Such a strategy works well when the auxiliary corpus is noisy as the fine-tuning step will only be trained on the gold annotations.

\textbf{Corpus Annealing ({\TEACHER})} Our last strategy was inspired by \newcite{clark-etal-2019-bam}'s teacher annealing proposal. \newcite{clark-etal-2019-bam} use teacher annealing to enable smoother learning. 
The multi-task model  initially learns from the predictions of the single-task model, but training gradually switches to gold labels by a weighted loss function. In this paper, we configured our system to  initially learn from the auxiliary corpus, and using a linearly decreasing ratio of training with the auxiliary corpus. 
Instead of using a weighted loss as 
done by %did in 
\newcite{clark-etal-2019-bam}, we used the ratio to control the source of our training documents (main or auxiliary).  By doing so, the learning process smoothly switches from the auxiliary corpus to the main corpus,
training 100\% on the main corpus when the end of  training is reached.

\subsection{Learning}
Following \newcite{lee2018higher}, we optimise our model on the marginal log-likelihood of all correct antecedents. 
We consider an antecedent correct if it is from the same gold single-antecedent coreference cluster \textsc{gold}$(i)$ as the gold antecedent lists. 
We also use the gold single-antecedent clusters to extend the split-antecedent anaphor list during training: i.e., mentions in the same single-antecedent cluster of a split-antecedent anaphor are considered as split-antecedent anaphors. The extension boosts the number of split-antecedent anaphors in the training data by 79\% to 908. %(see Table \ref{table:corpus_stat}). 
We compute the losses as follows:

\vspace{-8pt}
%\small
$$log \prod_{j=1}^{N}\sum_{\hat{y} \in Y(j) \cap \textsc{gold}(i)} r(\hat{y},j)$$
% \normalsize
in case mention $i$ is not a split-antecedent anaphor or $Y(j)$ (the candidate antecedents) does not contain mentions from $\textsc{gold}(i)$, we set \textsc{gold}$(i) = \{\epsilon\}$.

\section{Experiments}

\textbf{Datasets} We evaluated our models on the {\ARRAU} corpus \cite{poesio_anaphoric_2008,uryupina-et-al:NLEJ} as this is the only gold annotated corpus with split-antecedent anaphors annotated.\footnote{As far as we know the corpus used by \newcite{vala-etal-2016-antecedents} is not publically available, and the corpus used by \newcite{zhou&choi:COLING2018} only covers anaphoric references to a limited range of antecedents.} 
The corpus also contains annotations of the bridging references and single-antecedent coreference relations used for the auxiliary datasets. 
We used all four subcorpora of {\ARRAU}: \ACRO{rst} (news), \ACRO{trains} (dialogue),  \ACRO{pear} (fiction) and  \ACRO{gnome} (medical and art history). 
301 out of 552 total documents contain split-antecedent anaphors. 
We use the 1-7th, 8th, 9-10th of every 10 documents as our train, dev and test dataset, respectively (see Table \ref{table:corpus_stat} for more detail). 

In addition, we used the {\PDGAME} corpus to create auxiliary datasets.
%To form our auxiliary corpora, we additionally used the {\PDGAME} corpus {\PD}. 
The {\PD} corpus contains 542 documents from two main domains, Wikipedia and fiction. 
165 documents contain split-antecedent anaphors according to the silver labels in %comes with 
the corpus; we use those documents as our {\PDSILVER} corpus. 
Our {\PDCROWD} auxiliary corpus consists of the 467 documents which contain split antecedents when aggregated as %by the method 
described in Section \ref{sec:auxiliary}. 
The {\BRIDGING} corpus has 213 documents contain element-of bridging relations from the non-dev/test portion of the {\ARRAU} corpus. The {\COREF} corpus is formed by 462 non-dev/test documents of the {\ARRAU} corpus. 
Table \ref{table:corpus_stat} shows statistics about our corpora.

\textbf{Evaluation metrics} 
Following \newcite{vala-etal-2016-antecedents} we report \textbf{lenient} F1 scores that give partial credit 
when only some of the split individual antecedents of a plural are found,
%to the links between the split-antecedent anaphors and individual antecedents
and consider an antecedent correct as long as it belongs to the correct gold single-antecedent cluster. 
%For final evaluation, 
We further report  \textbf{strict} scores that require all antecedents of split-antecedent anaphors be correctly resolved for the final evaluation.

\textbf{Hyperparameters} We used  the default settings from \newcite{lee2018higher}, replacing %and replace 
their \ACRO{elmo} settings with the {\BERT} settings from \newcite{kantor-globerson-2019-coreference}. 
We trained the models (including pre-training models) for 200k steps.

\vspace{-4pt}
\section{Results and Discussions}

\vspace{-4pt}
\subsection{Training Strategy Selection}

We first applied all three training strategies to our auxiliary corpora to find the best training strategy for each auxiliary corpus. 
We used  lenient F1 scores on the development set to select the strategy most suitable for each individual corpus. 
As illustrated in Table \ref{table:tsselction}, our baseline model trained only on  {\ARRAU} {\ACRO{train}} already achieves a reasonably good F1 score for this task (58.2\%). Starting with a strong baseline, our system enhanced by the auxiliary corpora achieved substantial improvements of up to 11.3\%. 

Among the training strategies, 
%In term of the best training strategy for different auxiliary corpora,
{\PDSILVER} works best when using the {\CONCAT} method. 
This makes sense as  {\PDSILVER} contains split antecedent examples annotated using the same annotation scheme as {\ARRAU}. %the main training corpus. 
The {\PDCROWD} corpus is much noisier, but despite containing a large amount of false positives, it achieves better F1 scores than  {\PDSILVER}. 
This confirms our hypothesis that a higher recall of split-antecedent examples is important, and the false positive examples --mainly single-antecedent anaphors and bridging relations--do not harm the results. 
Both {\PRETRAIN} and {\TEACHER} are suitable strategies for the {\PDCROWD} corpus, with the former slightly better. 
The {\BRIDGING} corpus works best in a {\PRETRAIN} setting, with a large improvement of 6.2\% when compared to the baseline 
even though  
the corpus only contains a small number of examples (1k).
%it can boost the performance in our task by a large margin. 
The large improvement confirmed our hypothesis that \PRED{element-of} bridging relations are closely related to split-antecedent relations. 
% Although % Despite 
% the corpus only contains a small number of examples (1k), it can boost the performance in our task by a large margin. 
Finally, the {\COREF} corpus achieved the best scores with all three training strategy, but the largest improvement of 11.3\% is achieved by training with {\TEACHER} method. 
As the {\COREF} corpus has a substantially larger number of examples when compared to all other auxiliary corpus used in this paper, 
it seems likely that
%we suggest 
this might be an important reason for 
its usefulness.
%the successful use of the {\COREF} corpus. 
%We will make a further analysis in the later sections.
Overall, our auxiliary corpora and training strategies showed their merit for enhancing the performance on split-antecedent anaphora resolution;
we will further discuss this in  later sections.

\begin{table}[t]
\centering
%\small 
\resizebox{0.9\textwidth}!{
\begin{tabular}{lccccc}
\toprule
\bf Training Strategy&\bf \ACRO{baseline}&\bf {\PDSILVER}&\bf {\PDCROWD}&\bf {\BRIDGING}&\bf {\COREF}\\\midrule
{\CONCAT}   &58.2    &\bf 59.8  &61.2       &59.2       &67.6\\ 
{\PRETRAIN} &58.2    &59.0      &\bf 62.9   &\bf 64.3   &66.5\\
{\TEACHER}  &58.2    &59.0      &62.6       &61.1       &\bf 69.5\\ 
\bottomrule
\end{tabular}}
\caption{\label{table:tsselction} Training strategy selection on the development set with lenient F1 scores.}
\end{table}

\begin{table}[t]
\centering
%\small
\resizebox{0.75\textwidth}!{
\begin{tabular}{rccc|c}
\toprule
&\multicolumn{3}{c|}{\bf Lenient}&\bf Strict\\
&\bf R&\bf P&\bf F1&\bf Accuracy\\\midrule
\ACRO{recent-2}&19.6&21.8&20.6&3.6\\
\ACRO{recent-3}&31.8&23.6&27.1&0.9\\
\ACRO{recent-4}&40.4&22.6&28.9&0.0\\
\ACRO{recent-5}&45.7&20.4&28.2&0.0\\
\ACRO{random}&24.9&11.4&15.7&0.0\\%\midrule
\ACRO{neural baseline}&60.8&56.4&58.6&22.7\\\midrule
\PDSILVER&61.6&61.9&61.8&30.9\\
\PDCROWD&68.2&63.5&65.7&31.8\\
\BRIDGING&64.5&65.0&64.8&34.5\\
\COREF&68.6&\bf 70.6& 69.6& 42.7\\\midrule
{\PDCROWD} + \COREF&68.2&69.6&68.9&40.9\\
{\BRIDGING} + \COREF&69.4&67.5&68.4&39.1\\
%{\PDCROWD} + {\BRIDGING} + \COREF&66.5&68.5&67.5&41.8\\
{\PDCROWD} + {\BRIDGING} + \COREF&\bf 72.2&67.8&\bf 70.0&\bf 43.6\\
\bottomrule
\end{tabular}}
\caption{\label{table:final_results} 
Comparing our models with the baselines on the test set. 
}
\end{table}

\vspace{-4pt}
\subsection{Comparison with the Baselines}
We then evaluated our models, 
each trained using the best training strategy for that corpus, 
%with the different selected training strategies 
on the test set. 
Since our paper reports the first result on split-antecedent anaphora resolution on {\ARRAU}, we compare our system with various baselines. Following \newcite{vala-etal-2016-antecedents}, we created naive baselines: \ACRO{recent-m} and \ACRO{random}. 
\ACRO{recent-m} assigns to an anaphor  the  $m$ antecedents (from distinct single-antecedent clusters) that are closest. 
%to the anaphors. 
\ACRO{random} assigns all  candidate antecedents random probabilities;
the antecedents are selected using
the same method as the one used in our trained models. %to select antecedents for split-antecedent anaphors.

Table \ref{table:final_results} shows the results on the test set. 
The naive baselines achieved a maximum lenient F1 score of 28.9\% when using the $4$ most recent antecedents. 
When using strict evaluation, most  naive baselines 
perform really badly.
%fall to get any credit. 
This poor performance of the naive baselines confirms the difficulty of our task. 
Our {\ACRO{neural baseline}} trained solely on {\ARRAU} {\ACRO{train}} achieved a lenient F1 of 58.6\%, more than double  the best result of the naive baselines. 
With strict evaluation, the same model achieved 22.7\%--6 times  the best score of naive baselines, but still low.
Using  auxiliary corpora improved the performance of the neural model 
by a minimum of %further enhanced by at least 
3.2 and 8.2 p.p. when evaluated in a lenient and strict setting, respectively. But our best model, trained using  {\COREF} auxiliary information, achieved a lenient F1 of 69.6\% and a strict accuracy of 42.7\%, which   is 11\% and 20\% better than our {\ACRO{neural baseline}}.

We further evaluated using combinations of  auxiliary corpora--i.e., 
%We alternatively used 
using the {\PDCROWD}, {\BRIDGING}, and  {\COREF} corpora, 
and using either the {\PRETRAIN} or {\TEACHER} strategies--
%can be used as a pipeline, 
e.g., using  {\PDCROWD} for {\PRETRAIN},  %auxiliary corpus and 
then 
fine-tune the model with {\COREF} corpus using {\TEACHER} strategy. 
In total, we evaluated three combinations (see Table \ref{table:final_results}).
The best result, achieved by combining all three auxiliary corpora, was  0.4 p.p. and 0.9 p.p. better than the results achieved by using {\COREF} alone in a lenient and strict evaluation, respectively.
%although the result models outperform the {\PDCROWD} and {\BRIDGING} settings by large margins, the scores are still lower than the ones achieved by using {\COREF} alone. 

\subsection{Analysis}
\begin{table}[t]
\centering
%\small 
\begin{subtable}{\linewidth}
\centering
\resizebox{0.5\textwidth}!{
\begin{tabular}{lccccc}
\toprule
&&\multicolumn{2}{c}{\bf \ACRO{neural baseline}}&\multicolumn{2}{c}{\bf \ACRO{best}}\\
&\bf Count&\bf Lenient&\bf Strict &\bf Lenient&\bf Strict \\
\midrule
2   &157    &60.1  &33.1    &69.5   &44.0\\
3+  &33     &52.1  &6.1     &65.3   &21.2\\
\bottomrule
\end{tabular}}
\caption{\label{table:manaphra} Scores for anaphors with different number of antecedents.}
\end{subtable}
\begin{subtable}{\linewidth}
\centering
\resizebox{0.5\textwidth}!{
\begin{tabular}{lccccc}
\toprule
&\bf 1k&\bf 6k &\bf 10k &\bf 20k& \bf All (30k)\\\midrule
Lenient&60.1&63.3&65.9&67.0&69.5\\
Strict&28.8&32.5&40.0&46.3&46.3\\
\bottomrule
\end{tabular}}
\caption{\label{table:corpussize} Scores of models trained with reduced size of the {\COREF} corpus.}
\end{subtable}
\caption{Analysis of our best model.}
\end{table}

\begin{table}[t]
\small
\centering
\begin{tabularx}{\textwidth}{cX}
\toprule
\bf \ACRO{best}&The sudden romance of \CORRECT{British Aerospace} and \CORRECT{Thomson-CSF} -- traditionally bitter competitors for Middle East and Third World weapons contracts -- is stirring controversy in Western Europe 's defense industry . Most threatened by closer \ANAPHOR{British Aerospace-Thomson} ties would be their respective national rivals\\\midrule
\bf \ACRO{baseline}&The sudden romance of \CORRECT{British Aerospace} and \MISSING{Thomson-CSF} -- traditionally bitter competitors for Middle East and \WRONG{Third World weapons contracts} -- is stirring controversy in Western Europe 's defense industry . Most threatened by closer \ANAPHOR{British Aerospace-Thomson} ties would be their respective national rivals\\\midrule\midrule

\bf \ACRO{best}&Workers dumped large burlap sacks of \CORRECT{the imported material} into a huge bin , poured in cotton and \CORRECT{acetate fibers} and mechanically mixed \ANAPHOR{the dry fibers} in a process used to make filters .\\\midrule
\bf \ACRO{baseline}&Workers dumped large burlap sacks of \MISSING{the imported material} into a huge bin , poured in cotton and \CORRECT{\WRONG{acetate} fibers} and mechanically mixed \ANAPHOR{the dry fibers} in a process used to make filters .\\\midrule\midrule

\bf \ACRO{best}&\CORRECT{Time Warner Inc.} is considering a legal challenge to \CORRECT{Tele-Communications Inc.} 's plan to buy half of Showtime Networks Inc. , a move that could lead to all-out war between \ANAPHOR{the cable} \ANAPHOR{industry 's two most powerful players} . \\\midrule
\bf \ACRO{baseline}&\MISSING{Time Warner Inc.} is considering a legal challenge to \WRONG{\MISSING{Tele-Communications Inc.} 's plan to} \WRONG{buy half of \NESTWRONG{Showtime Networks Inc.}} , a move that could lead to all-out war between \ANAPHOR{the cable} \ANAPHOR{industry 's two most powerful players} .\\\midrule\midrule

\bf \ACRO{best}&In \CORRECT{California} and \CORRECT{New York} , state officials have opposed Channel One . Mr. Whittle said private and \WRONG{parochial \NESTWRONG{schools}} in \ANAPHOR{both states} will be canvassed to see if they are interested in ...\\\midrule
\bf \ACRO{baseline}&In \MISSING{California} and \MISSING{New York} , state officials have opposed Channel One . Mr. Whittle said private and \WRONG{parochial \NESTWRONG{schools}} in \ANAPHOR{both states} will be canvassed to see if they are interested in ...\\
\bottomrule
\end{tabularx}
\caption{\label{table:example} 
A comparison of system prediction examples from our \ACRO{best} and \ACRO{baseline} system. 
The colours indicate the correctness of the predicted split-antecedents 
%are colourfully marked 
(\CORRECT{true positive}, \MISSING{false negative}, \WRONG{false positive});
the \ANAPHOR{anaphors} are marked with underlines.
}
\end{table}

\textbf{Number of Antecedents} 
%We further group the split-antecedent anaphors by the number of antecedents they have and compare the lenient and strict scores between our best model and the baseline. 
We compared our best model with the neural baseline, using both lenient and strict scores and also considering the number of split-antecedents. 
For this analysis, we evaluated both models on the concatenation of test and development set to collect more examples. 

As shown in Table \ref{table:manaphra}, 82.6\% of the anaphors in the dataset have two antecedents; the rest have three or more antecedents. 
With anaphors that have two antecedents, our best model achieved improvements of  9.5\% (lenient) and 10.9\% (strict), respectively.
%on the lenient and strict scores respectively. 
With anaphors %For the group 
that have more antecedents, our best model achieved even larger improvements for both lenient (13.2\%) and strict (15.1\%) scores. 
Overall, our best model outperforms the baseline by large margins in all evaluations. System prediction examples for both systems can be find in Table \ref{table:example}. %Appendix \ref{appendix:example}. 
%our best model improves the lenient score by 6.7\%, the large gain is as a result of a better precision. In terms of the strict evaluation, apart from the 6\% of the anahpors that have been fully resolved, both systems have correctly recalled at least 2 antecedents for another 51.5\% anaphors. For those anaphors, the final system generates way less false positive examples than the baseline. E.g. in ``Bowing to criticism , \textit{Bear Stearns}$_i$ , \textit{Morgan Stanley}$_j$ and \textit{Oppenheimer}$_k$ joined PaineWebber in suspending stock-index arbitrage trading for \underline{their}$_{i,j,k}$ own accounts.", our final system successfully recalled \textit{Morgan Stanley}and \textit{Oppenheimer} and the baseline assigns incorrectly the \textit{PaineWebber} as an addition antecedent of \underline{their}. Overall, our best system achieved large improvements for both groups on lenient scores. The improvements on the strict scores are as a result of better predictions on anaphors with two antecedents.

\textbf{Size of the Auxiliary Corpus} Our {\COREF} auxiliary corpus achieved much larger improvements than all other corpora evaluated in this paper. 
A simple explanation would be that this is because the {\COREF} corpus is substantially larger. % (in term of examples). 
So, to understand the impact of the auxiliary corpus size on our task, we further trained our model with auxiliary corpora of different size. 
The examples are randomly selected from our {\COREF} corpus. 
Table \ref{table:corpussize} shows our results on the development set. When using 1k examples from the {\COREF} corpus, the lenient F1 is 4.2\% lower than {\BRIDGING}'s 64.3\% which suggest the {\BRIDGING} corpus is more effective when the number of training examples is similar. 
When compared with  {\PDCROWD}, the same amount of the gold annotated single-antecedent coreference examples (6k) achieved  broadly the same score. Adding more training examples results in an increase in lenient scores. The strict scores follow a similar trend up until 2/3 of the examples are used (20k). Overall, the auxiliary corpus size is an important factor for the final results. 

\vspace{-4pt}
\section{Other Approaches to Split-Antecedent Anaphora Resolution}
\label{section:relatedwork}
Recently, \newcite{vala-etal-2016-antecedents} introduced the first modern system to resolve split antecedent anaphora, 
although focusing only on  plural pronouns \LINGEX{they} and \LINGEX{them}, 
%that attempted to resolve two specific split-antecedent anaphors (\LINGEX{they} and \LINGEX{them}) 
and using a  corpus  of fiction they themselves annotated.
%annotated in house. 
\newcite{vala-etal-2016-antecedents} proposed a learning-based system using handcrafted features,
%to resolve only they/them anaphors find in the corpus. 
which achieved %They report 
a score of 43.4\% using 
%the same  method of our 
the lenient evaluation they proposed 
and 
we adopted. % this evaluation method.  
%We argue that our 
The version of the task tackled in this paper is harder in 
three 
respects.
%task is harder than theirs, firstly, 
First, our system resolves all split-antecedent references, without restriction.
Secondly, %evaluated their system on fiction domain we evaluate 
our system was evaluated on the full {\ARRAU} corpus \cite{uryupina-et-al:NLEJ}, which contains text from multiple genres (news, dialogues, stories, medical and art history). 
% In the fiction genre, the antecedents of \LINGEX{they} / \LINGEX{them} 
% %form a fiction domain text 
% are mostly from a limited %a countable 
% list of characters in the novel, 
% which makes the task  easier.
Thirdly, in addition to Vala \textit{et al.}'s lenient evaluation that gives partial credit to split-antecedent anaphors not all of whose antecedents are identified,
%that are not resolved fully correct, 
we also report  strict scores that only give credit to the model when all the antecedents of an anaphor are correctly resolved.

More recently, \newcite{zhou&choi:COLING2018} introduced a corpus for entity linking and coreference   in  transcripts of the \textit{Friends} sitcom.
Plural mentions are annotated if they are linked to the main characters; as a result, the vast majority (95\%) of the plurals in this corpus are pronouns. 
And since the corpus was primarily created for entity linking, its plural annotations are problematic for coreference: 58.8\% of  plural mentions are linked  either to \textit{General} entities that are not annotated in the text, 
or to characters that do not appear in the utterance before the plural anaphor. 
Also, 
only the results 
%the proposed neural system %based on CNNs 
%was only evaluated  
on the 
combination of singular and plural mentions
are reported;
%coreference clusters mix of both singular and plural mentions,
the performance  on plural mentions only is not.  

\vspace{-4pt}
\section{Conclusions}
We propose the first model for unrestricted split-antecedent anaphora resolution. 
Starting %with a strong baseline adapted 
from the SoTA single-antecedent coreference resolution system,
we substantially improve its performance on the task through a combination 
of exploiting auxiliary corpora for related tasks.
%and further enhance the performance of the task by using four different auxiliary corpora with 
%using according to several  training strategies. 
Despite our baseline having a good performance of 58.6\% (lenient F1), our best model achieved large gains of 11 percentage points. Further, evaluation using strict accuracy shows our best system can correctly resolve 43.6\% of  split-antecedent anaphors, which is 21 p.p. better than our baseline. 
%Overall, our system showed it merit on resolving the split-antecedent anaphors. 

\section*{Acknowledgements}
This research was supported in part by the DALI project, ERC Grant 695662.

%The acknowledgements should go immediately before the references.  Do not number the acknowledgements section. Do not include this section when submitting your paper for review.

\bibliographystyle{coling}
\bibliography{coling2020}
\end{document}